\documentclass[letterpaper]{article}
\usepackage{aaai}
\usepackage{times}
\usepackage{helvet}
\usepackage{courier}
\usepackage{booktabs}
\usepackage{graphicx}
\usepackage{subcaption}
\usepackage{amsmath}
\usepackage{xcolor}
\usepackage{latexsym}
\usepackage{tikz}
\usepackage{etoolbox}
\usepackage{booktabs}
\usepackage{natbib}
\usepackage{tikzsymbols}
\usepackage{bm}
\usepackage{mathtools}  % amsmath with extensions
\usepackage{amsfonts}  % (otherwise \mathbb does nothing)
\usepackage{package_zhijing}

\usepackage{adjustbox}
\usepackage{verbatim}
\usepackage{caption}
\usepackage{multirow}
\begin{document}
% The file aaai.sty is the style file for AAAI Pr ess 
\title{
% Cross-Attention Contrastive Learning for \\ 
Inconsistent Few-Shot Relation Classification
% Generalized Few-Shot Relation Classification
via \\
Cross-Attentional Prototype Networks with Contrastive Learning
}
\author{
Hongru Wang $^{\star}$,
Zhijing Jin $^{\clubsuit}$,
Jiarun Cao $^{\ddagger}$,
Gabriel Pui Cheong Fung $^{\star}$,
Kam-Fai Wong $^{\star}{\dagger}$ \\
 $^{\star}$ Department of Systems Engineering and Engineering Management, The Chinese University of Hong Kong \\
$^{\clubsuit}$  Max Planck Institute for Intelligent Systems \& ETH Zürich  \\
% University of Manchester \\
 \{hrwang, kfwong\}@se.cuhk.edu.hk
}

\maketitle
\begin{abstract}
\begin{quote}
Standard few-shot relation classification (RC) is designed to learn a robust classifier with only few labeled data for each class. However, previous works rarely investigate the effects of a different number of classes (i.e., $N$-way) and number of labeled data per class (i.e., $K$-shot) during training vs. testing. In this work, we define a new task, \textit{inconsistent few-shot RC}, where the model needs to handle the inconsistency of $N$ and $K$ between training and testing. To address this new task, we propose Prototype Network-based cross-attention contrastive learning (ProtoCACL) to capture the rich mutual interactions between the support set and query set. Experimental results demonstrate that our ProtoCACL can 
% (1) achieve better performance with less data under inconsistent few-shot setting than standard few-shot setting, and (2)
outperform the state-of-the-art baseline model under both inconsistent $K$ and inconsistent $N$ settings, owing to its more robust and discriminate representations. Moreover, we identify that in the inconsistent few-shot learning setting, models can achieve better performance with \textit{less data} than the standard few-shot setting with carefully-selected $N$ and $K$. In the end of the paper, we provide further analyses and suggestions to systematically guide the selection of $N$ and $K$ under different scenarios.
% at the last.

\footnote{$^{\dagger}$ Corresponding Author}
 
\begin{comment}
Few-shot RC is a task of predicting the relationship that exists in the given sentence from few labeled samples, which is formulated as $N$-way $K$-shot problem in previous works. This setting strictly requires consistent $N$ and $K$ during training and testing. However, the limitation is hard to hold in practice due to data sparsity, especially for target data. \textbf{Inconsistent few-shot RC}, which frees this constraint and does not require the consistent $N$ and $K$ during training and testing. In this paper, we firstly define the inconsistent few-shot problem and investigate the performance of the state-of-the-art model under this setting. Besides, 
\end{comment}

% In this paper, we firstly evaluate the performance of ProtoNets under inconsistent few-shot setting, and then we proposed cross attention contrastive learning framework based on our observations. Experimental results indicates 1) CACL outperform Proto in both inconsistent $K$ and inconsistent $N$ setting, 2) inconsistent setting can outperform standard setting under different scenarios, especially with less data required. We left inconsistent few-shot at token-level in our future work.
\end{quote}
\end{abstract}

\section{Introduction}
Relation classification (RC) is an important task in natural language processing which aims to capture the relation of two entities in the given text. Most supervised models require large and high-quality annotated data \citep{zelenko2003kernel,surdeanu2012multi,zeng2014relation,gormley2015improved,shi2019simple}. However, in practice, it is often difficult to acquire large volumes of labeled data, which leads to poor performance of supervised learning models. To tackle this data scarcity problem, researchers propose the task of few-shot RC, to use only few labeled examples to learn the RC task \citep{han-etal-2018-fewrel,gao-etal-2019-fewrel,sun-etal-2019-hierarchical,qu2020few}. 

\begin{table}[t]
\centering
\small
\resizebox{\columnwidth}{!}{%
\begin{tabular}{l@{\hspace{0.6\tabcolsep}}l@{\hspace{0.6\tabcolsep}}p{0.68\columnwidth}}
\toprule
\multicolumn{3}{c}{\textbf{Training Phase }}\\
\midrule
\multirow{13}{*}{\textbf{Supp. Set}} & \multirow{2}{*}{(A) capital\_of}   & 
 \emph{\textcolor{blue}{London}} is the capital of \emph{\textcolor{red}{the U.K}}.  \\
\cline{3-3}
& & \emph{\textcolor{blue}{Washington}} is the capital of \emph{\textcolor{red}{the U.S.A}}. \\
\cline{2-3}
% \cmidrule(lr){2-3}
& \multirow{4}{*}{(B) member\_of} & \emph{\textcolor{blue}{Newton}} served as the president of \emph{\textcolor{red}{the Royal Society}}. \\
\cline{3-3}
& & \emph{\textcolor{blue}{Leibniz}} was a member of \emph{\textcolor{red}{the Prussian Academy of Sciences}}. \\
% \cmidrule(lr){2-3}
\cline{2-3}
& \multirow{6}{*}{ (C) birth\_name }    & \emph{\textcolor{red}{Samuel Langhorne Clemens}}, better known by his pen name \emph{\textcolor{blue}{Mark Twain}}, was an American writer. \\
\cline{3-3}
&  & \emph{\textcolor{blue}{Alexei Maximovich Peshkov}}, primarily known as \emph{\textcolor{red}{Maxim Gorky}}, was a Russian and Soviet writer. \\
\midrule
\multirow{2}{*}{\textbf{Query}}  &   \multirow{2}{*}{(A), (B) or (C) }     &  \emph{\textcolor{blue}{Euler}} was elected a foreign member of \emph{\textcolor{red}{the Royal Swedish Academy of Sciences}}. \\
\midrule
\midrule
\multicolumn{3}{c}{\textbf{Test Phase}}\\
\midrule
\multirow{3}{*}{\textbf{Supp. Set}} & \multirow{1}{*}{(A) date\_of\_birth}     & \emph{\textcolor{blue}{Mark Twain}} was born in \emph{\textcolor{red}{1835}}. \\
% \cmidrule(lr){2-3}
\cline{2-3}
& \multirow{2}{*}{(B) place\_of\_birth}    & \emph{\textcolor{blue}{Elvis Presley}} was born in \emph{\textcolor{red}{Memphis, Tennessee}}. \\
\midrule
\multirow{2}{*}{\textbf{Query}}      &   \multirow{2}{*}{ (A) or (B) }      &  \emph{\textcolor{blue}{William Shakespeare}} passed away at age 52 (around \emph{\textcolor{red}{1616}}).\\

\bottomrule
\end{tabular}
}
\caption{An example for a 2-way 1-shot scenario at testing and 3-way 2-shot at training, including both inconsistent $K$ few-shot and inconsistent $N$ few-shot. Different colors indicate different entities, \textcolor{red}{red} for tail entities, and \textcolor{blue}{blue} for head entities. For inconsistent $K$ few-shot problem, the number of shots is not inconsistent between training and testing. For inconsistent $N$ few-shot problem, the number of labels is not inconsistent during training and testing.}
\label{tab:dataformat}
\end{table}

% 第一个后果就是 每一个new class都需要重新set如果是不同的k
% 第二个后果就是
Typically, standard few-shot RC is formulated as follows: given a set of base classes $C_{\mathrm{base}}$ with many labeled samples and a set of novel classes $C_{\mathrm{novel}}$ with only $K$ samples per class, we aim to classify the relation of query instances to a type in the set $C_{\mathrm{novel}}$. Most existing works address this problem by a meta-learner to extract information from base classes during meta-training and then \textit{directly} adapt to novel classes during meta-testing \citep{han-etal-2018-fewrel,gao-etal-2019-fewrel,sun-etal-2019-hierarchical}. However, one constraint in the existing approaches is that they require both the training and testing data have the \textit{same} number of classes (often denoted as $N$-way) and number of samples each class (often denoted as $K$-shot). Specifically, in both the training and test sets, there are many subtasks, each of which has a support set and a query set. The query set contains only unlabeled test instances, and the support set includes $N$ class and $K$ annotated samples for each class for both the base and novel classes.
% both base and novel classes need to be $N$-way $K$-shot, in which each sub-task contains a support set and a query set. The support set includes $N$ class and $K$ annotated samples for each class and the query set contains a set of unlabeled test instances. 
However, this constraint might not be suitable for real scenarios. First, it is hard to hold this consistent $N$-way $K$-shot setting in practice, because the numbers of base classes and novel classes can vary as more data is collected. Also, it is fortuitous and implausible to have precisely the same amount of data for novel classes. For example, there can be $m$ labeled examples for relation A but $n$ labeled examples for relation B where $m \ne n$. Second, it is not guaranteed that this consistency brings best performance. \cite{gao2019hybrid} prove that feeding more classes during training led to better performance, and, one step further, \cite{cao2019theoretical} provide a theoretical analysis for inconsistent $K$ problem for few-shot image classification problem. 

%The model should be able to show robust generalization capability without retraining and much performance diminishment.

In this work, we propose a more realistic and challenging task in low-resource scenarios: \textit{inconsistent few-shot RC}, which can be further divided into inconsistent $N$ and inconsistent $K$ problems as shown in Table~\ref{tab:dataformat}. Different from the standard few-shot problem, our proposed inconsistent few-shot task does not require consistent $N$ and $K$ during training and testing, which is challenging but more common in practice. Inconsistent $N$ means that the numbers of classes are not consistent during training and testing while inconsistent $K$ means the numbers of shots are not consistent. Unfortunately, many models for solving the standard few-shot RC problem are incompatible with our inconsistent few-shot RC setting \citep{gao2019hybrid,sun-etal-2019-hierarchical,qu2020few}. Few models such as 
% the Matching Network \citep{vinyals2016matching}, and a more recent, better performing model, 
Prototypical Networks (ProtoNets) \citep{NIPS2017_cb8da676}
% \citep{NIPS2017_cb8da676,vinyals2016matching}
can be directly used to solve this problem, but the performance is substantially affected by inconsistent $N$ and $K$.

To address these challenges, we propose a novel model, ProtoNet-based cross-attention and contrastive learning (ProtoCACL). Our model is motivated by the problem that previous studies either implicitly model the support set and query set with a shared encoder or only explicitly consider the single information flow from support set to query set \citep{gao2019hybrid}. None of these approaches models the bidirectional interactions between these two sets. Instead of modeling these sets separately, we propose a cross-attention module to capture the relationship between the support set and query set. In addition, we utilize contrastive learning to add additional constraints for both support set and query set. 

In summary, the contributions of our paper are as follows: (1) We propose a new task named Inconsistent Few-shot RC, which covers inconsistent $N$ and inconsistent $K$ settings. (2)  We propose a novel model ProtoCACL to tackle this new task, by learning more robust and discriminative representations. (3) We conduct extensive experiments, under various inconsistent few-shot RC settings such as inconsistent $K$ and inconsistent $N$, and verify that our method is more robust and outperforms the strongest baseline by up to 1.41\%. Moreover, inconsistent few-shot RC can achieve better performance than the standard few-shot RC by using carefully selected $N$ and $K$. We provide some suggestions to guide the selection of $N$ and $K$ for this problem.

\section{Related Work}
% maml proto mn
Few-shot learning (FSL) aims to acquire and transfer knowledge from few labeled samples (i.e., support set). 
Transfer methods are first proposed using a Bayesian approach on low-level visual features \citep{1597116}. Then, metric learning methods \citep{vinyals2016matching,NIPS2017_cb8da676} were proposed to learn different representations across classes. One representative work is prototypical networks \citep{NIPS2017_cb8da676}, aiming to learn robust class representations and classify the query set based on the distance between class representation and sample vector in the feature space. Another line of work is optimization-based method \citep{ravi2016optimization,finn2017model,nichol2018first}. For example, model agnostic meta-learning (MAML) \citep{finn2017model} uses a few gradient update steps computed with a small amount of data from new task to get proper parameters from the initial parameter space, and Reptile \citep{nichol2018first} simplify the implementation of MAML by a first-order optimization-based meta-learning method.

% cite coach
FSL has been applied to many NLP tasks such as slot tagging \citep{lee2018zeroshot,hou2020fewshot,liu-etal-2020-coach,wang2021mcml}, text generation \citep{mi2019meta,peng2020few}, and also RC \citep{han-etal-2018-fewrel,gao2019hybrid,qu2020few}. For standard few-shot RC, \cite{han-etal-2018-fewrel} present the first dataset FewRel with 70K sentences on 100 relations derived from Wikipedia. FewRel 2.0 \citep{gao-etal-2019-fewrel} extends the dataset with few-shot domain adaption and few-shot none-of-the-above detection. Many works on FewRel datasets focus on improvement of methods, including modeling the distance distribution \citep{gao2019hybrid,ding2021prototypical}, utilizing external knowledge such as knowledge graph \citep{qu2020few}, learning different level of features \citep{sun-etal-2019-hierarchical,ye-ling-2019-multi}, and using pre-trained language models \citep{soares2019matching}. Apart from method innovation on the standard setting of consistent few-shot RC, the investigation for inconsistent few-shot RC still in demand. 

\begin{comment}
All these methods fall into the standard few-shot setting which requires consistent $N$ and $K$ during training and testing. However, it is fortuitous and implausible to have precisely the same quantity of data for the target and source tasks in practice, especially under few-shot scenario. There are few works that focus on inconsistent few-shot problems. \cite{cao2019theoretical} introduce a solid theoretical analysis of the impact of the shot number on ProtoNets, which explain the performance variation under inconsistent $K$ and validate the observation through experiments on few-shot image classification task. More efforts to solve inconsistent few-shot text classification are still in demand.
\end{comment}

% check the related work of this paper
Our paper considers RC in an inconsistent FSL scenario, which is a novel perspective that is under-explored
% , which is undermined 
in previous works. Due to the existence of inconsistent $N$ and K, most prior works cannot be used directly in this case, and the very few models that can be applied to this problem shows
% the few works that have been used to address this problem have been significantly influenced, resulting in 
relatively poor performance. More specifically, we investigate the effects of varying number of classes and number of shots under inconsistent FSL scenario, and experimental result shows that inconsistent few-shot reaches higher performance than standard few-shot setting with fewer data required (see next section). In addition, we provides an analysis and guidelines about the selection of $N$ and $K$ systematically which is our main contribution. All of these reflect the importance and value of research for inconsistent few-shot problem that motivate our work.

\section{Problem Formulation}

In this section, we first introduce the task formulation of inconsistent few-shot RC, and then present three challenges of this task by analyzing the limitations of the commonly used method, ProtoNet, under this new setting.
% Furthermore, we presents performance of ProtoNet under this setting and provide a detailed analysis.

\subsection{Standard Few-Shot Relation Classification}
Few-shot RC aims to predict the relation between two entities (i.e., head and tail entity) in a sentence, which is formulated as $N$-way $K$-shot problem \citep{gao2019hybrid,han-etal-2018-fewrel}. 

We denote a relation set consisting of $N$ relations as $\mathbb{R} = \{r_1, r_2, \dots, r_N \}$. Each sample $\bm{s} := (\bm{x}, \bm{h}, \bm{t})$ contains a sentence $\bm{x}$ with two annotated entities, i.e., a head entity $\bm{h}$ and a tail entity $\bm{t}$. For example, the sentence $\bm{x}=$ ``\emph{\textcolor{blue}{London}} is the capital of \emph{\textcolor{red}{the U.K.}}'' has a head entity $\bm{h}=$ ``\emph{{London}}'' and a tail entity $\bm{t}=$ ``\emph{{the U.K}}.''

There are two sets of samples: the support set $\mathbb{S}$ contains all labeled samples, namely samples with their labeled relation types, i.e., $(\bm{s}, r)$, where $r \in \mathbb{R}$, while the query set $\mathbb{Q}$ contains all unlabeled samples, namely samples that do not have the labels of their corresponding relation types \footnote{We have labels during training}. 

The models need to learn from the support set and predict the relation for the given samples in the query set.

In the standard few-shot RC setting,
% $N$-way $K$-shot RC setting, the goal is to learn the missing relations of samples in the query set $\mathbb{Q}$ by the labeled relations in the given support set $\mathbb{S}$.
the support set $\mathbb{S}$ contains $N \times K$ labeled samples, where $K$ is the number of samples for each of the $N$ relation types. Note that all previous works have a strict assumption that each relation type $r \in \mathbb{R}$ must have (1) {the same number ($K$) of} labeled samples, and (2) the same number ($N$) of relation types in the support set $\mathbb{S}$, across \textit{all training, validation and test sets}.

% In standard few-shot setting, all data in train, validation and test must strictly comply $N$-way $K$-shot as follows,

% \begin{equation}
%     N = M = |R|, K = n_1 = \dots = n_N
% \end{equation}

\subsection{Inconsistent Few-Shot Relation Classification}
We propose a new setting, inconsistent few-shot RC, to relax the strict assumption that all the support sets in training, validation and test sets should have the same number ($K$) of labeled samples across $N$ relation types. Instead, we allow a general setting where the support set $\mathbb{S}_{\mathrm{train}}$ of the training set can have $N_1$ relation types, and $K_1$ labeled samples per relation, while the support set $\mathbb{S}_{\mathrm{infer}}$ in the inference stage (i.e., in the validation and test sets) can have $N_2$ relation types, and $K_2$ labeled samples per relation, namely the training set:
\begin{align}
\small
\mathbb{R}_{\mathrm{train}} = \{ & r_1, r_2, \dots, r_{N1}\}
\\
\begin{split}
% \mathbb{S}_{\mathrm{train}} = \{ (\{\bm{s}_i^j\}_{j=1}^{K_1}, r_i) \}_{i=1}^{N_1}
% \\
\mathbb{S}_{\mathrm{train}} = \{ & (\bm{s}_r^1, r), \dots, (\bm{s}_r^{K_1}, r) | \text{ for each } r \in \mathbb{R}_{\mathrm{train}} \}
% \\
% \mathbb{S}_{\mathrm{train}} = \{ & (\bm{s}_1^1, r_1), \dots, (\bm{s}_1^{K_1}, r_1), \\
%     & \dots, \\
%     & (\bm{s}_i^1, r_j), \dots, (\bm{s}_i^{K_1}, r_j), \\
%     & \dots, \\
%     & (\bm{s}_{N_1}^1, r_{N_1}), \dots, (\bm{s}_{N_1}^{K_1}, r_{N_1}) \}
    ~,
\end{split}
\end{align}
and the inference set (i.e., validation and test sets):
\begin{align}
\small
\mathbb{R}_{\mathrm{infer}} = \{ & r_1, r_2, \dots, r_{N2}\}
\\
\begin{split}
\mathbb{S}_{\mathrm{infer}} = \{ & (\bm{s}_r^1, r), \dots, (\bm{s}_r^{K_2}, r) | \text{ for each } r \in \mathbb{R}_{\mathrm{infer}} \}
% \mathbb{S}_{\mathrm{test}} = \{ & (\bm{s}_1^1, r_1), \dots, (\bm{s}_1^{K_2},     r_1), \\
%     & \dots, \\
%     & (\bm{s}_i^1, r_j), \dots, (\bm{s}_i^{K_2},     r_j), \\
%     & \dots, \\
%     & (\bm{s}_{N_2}^1, r_{N_2}), \dots, (\bm{s}_N^{K_2}, r_{N_2}) \}
    ~.
\end{split}
\end{align}

This more general setup can encourage better flexibility and adaptability of models to real-world scenarios, in cases where $K_1 \ne K_2$ (the problem of inconsistent $K$) or $N_1 \ne N_2$ (the problem of inconsistent $N$). 
% In this paper, we target inconsistent few-shot problem based on two observations: 1) inconsistent few-shot problem is more common in practice and standard setting is a unique rationale of inconsistent setting, and 2) As illustrated in our pilot experiment, additional data may lead to performance degradation and unnecessary cost.

\subsection{Challenges of Inconsistent Few-Shot RC}
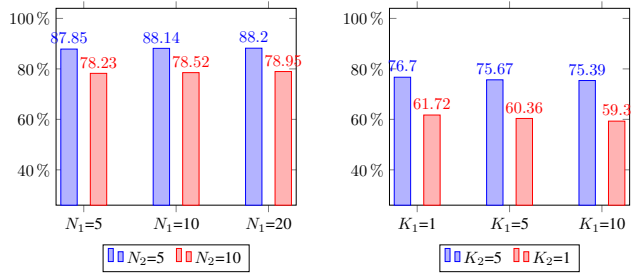
\begin{figure}
\begin{tikzpicture}[scale=0.62]
\begin{axis}[
    x tick label style={
        /pgf/number format/1000 sep=},
    xtick={0,1,2},
    xticklabels={$N_1$=5,$N_1$=10,$N_1$=20},
    % ylabel=Accuracy,
    yticklabel=\pgfmathprintnumber{\tick}\,$\%$,
    enlargelimits=0.15,
    legend style={at={(0.5,-0.2)},
        anchor=north,legend columns=-1},
    ybar=8pt,% configures `bar shift'
    bar width=10pt,
    nodes near coords,
    ymin=35,
    ymax=95,
    width=0.8\columnwidth
]
\addplot coordinates {(0,87.85)(1,88.14) (2,88.20)};
\addplot coordinates {(0,78.23)(1,78.52) (2,78.95)};
\legend{$N_2$=5 { } ,$N_2$=10}
\end{axis}
\end{tikzpicture}
\hfill
\begin{tikzpicture}[scale=0.62]
\begin{axis}[
    % x tick label style={
        % /pgf/number format/1000 sep=},
    xtick={0,1,2},
    xticklabels={$K_1$=1,$K_1$=5,$K_1$=10},
    % ylabel=Accuracy,
    yticklabel=\pgfmathprintnumber{\tick}\,$\%$,
    enlargelimits=0.15,
    legend style={at={(0.5,-0.2)},
        anchor=north,legend columns=-1},
    ybar=8pt,% configures `bar shift'
    bar width=10pt,
    nodes near coords,
    ymin=35,
    ymax=95,
    width=0.8\columnwidth
]
\addplot coordinates {(0,76.70)(1,75.67) (2,75.39)};
\addplot coordinates {(0,61.72)(1,60.36) (2,59.30)};
\legend{$K_2$=5 { } ,$K_2$=1}
\end{axis}
\end{tikzpicture}
\caption{Accuracy of ProtoNet on FewRel 1.0 validation dataset.
\textbf{Left:} inconsistent $N$ few-shot problem with $K$ set as 5 during training and testing (trainN-testN-5-5). \textbf{Right:} inconsistent $K$ few-shot problem with
N set as 5 during training and testing (5-5-trainK-testK).}
\label{fig:initial_exp}
\end{figure}
% \begin{figure}[t]
% 	\centering
% 	\footnotesize
% 	\begin{tikzpicture}
% 	\draw (0,0) node[inner sep=0] {\includegraphics[width=1.0\columnwidth, height=4cm, trim={0cm 0cm 0cm 0cm}, clip]{figures/inconsistent_n_proto.pdf}};
% 	\draw (0,-4) node[inner sep=0] {\includegraphics[width=1.0\columnwidth, height=4cm, trim={0cm 0cm 0cm 0cm}, clip]{figures/inconsistent_k_proto.pdf}};
% 	\end{tikzpicture}
% 	\caption{ Accuracy of ProtoNet on FewRel 1.0 validation dataset. \textbf{Upper:} inconsistent $K$ few-shot problem with $N$ set as 5 during training and testing (5-5-trainK-testK); \textbf{Lower:} inconsistent $N$ few-shot problem with $K$ set as 5 during training and testing (trainN-testN-5-5).
% 	}\vspace*{-4mm}
% 	\label{fig:initial_exp}
% \end{figure}

We analyze the challenges brought by this inconsistent few-shot RC setting between the training stage and the inference stage in terms of $N$ (the number of relation types) and $K$ (the number of samples per relation in the support set).

Specifically, we investigate the performance of one of the most common models, ProtoNets \citep{NIPS2017_cb8da676}, under this inconsistent few-shot RC setting. As shown in Figure~\ref{fig:initial_exp}, with the same training set setting $N_1$ or $K_1$, the model performance at the inference stage varies a lot by the inference set setting $N_2$ or $K_2$. For example, when there are 5 relation types in the training set ($N_1=5$), the trained model can have 87.85\% accuracy if the inference set has 5 relation types ($N_2=5$), but drops to 78.23\% accuracy if the inference set has 10 relation types ($N_2=10$), which is a substantial performance difference.

% t is obvious that the performance will be dramatically influenced by the inconsistent $N$ and $K$. 
There are three main findings from this analysis:
% Furthermore, some interesting and important conclusions can be drawn from this initial experiment: 

(1) Under the \textit{same training setting} ($N_1$ or $K_1$), \textit{more challenging inference settings} (larger $N_2$ or smaller $K_2$)
% When the training shots (or training ways) keep the same, less testing shots (or more testing ways) 
can \textit{seriously affect} the performance. This has also been partly observed by \citet{cao2019theoretical} in few-shot image classification. In Figure~\ref{fig:initial_exp}, such performance drop can be up to 16.09\%, in the setting of $K_1=10, K_2=5 \text{ or }1$. % For example, when the number of shots during training is 5, the accuracy of ProtoNets drops from 75.67\% to 60.36\% with the number of shots during testing as 5 and 1 respectively. 

(2) Under the \textit{same inference setting} ($N_2$ or $K_2$), \textit{easier training settings} (smaller $N_1$ or larger $K_1$) can lead to \textit{slightly poorer performance} during the inference time.
% \textbf{When the testing shots (or testing ways) are invariant, more training shots (or less training ways) damage the performance,}
In our experiments with ProtoNet, this drop is up to 2.42\%, in the setting of $K_2=1, K_1=1$ \text{ or }10. This observation also echoes with the analysis by \citet{gao2019hybrid}.

(3) \textit{Inconsistent} few-shot RC can reach \textit{better performance} than standard few-shot problem with carefully designed settings. For instance, the inconsistent setting of $N_1$-$N_2$-$K_1$-$K_2$=5-5-1-5 can get higher accuracy than the standard setting of $N_1$-$N_2$-$K_1$-$K_2$=5-5-5-5 (76.70\% vs 75.67\%). More interestingly, computational resources can be saved largely under this setting. Such evidence indicates the potentially large importance and value of research for inconsistent few-shot problem. 
% More detailed analysis is presented in Section 6.

\begin{figure*}[t]
	\centering
	\footnotesize
	\begin{tikzpicture}
	
	\begin{subfigure}[t]{1.0\columnwidth}
	    \draw (0,0) node[inner sep=0] {\includegraphics[width=1.0\columnwidth, height=6cm, trim={0cm 0cm 0cm 0cm}, clip]{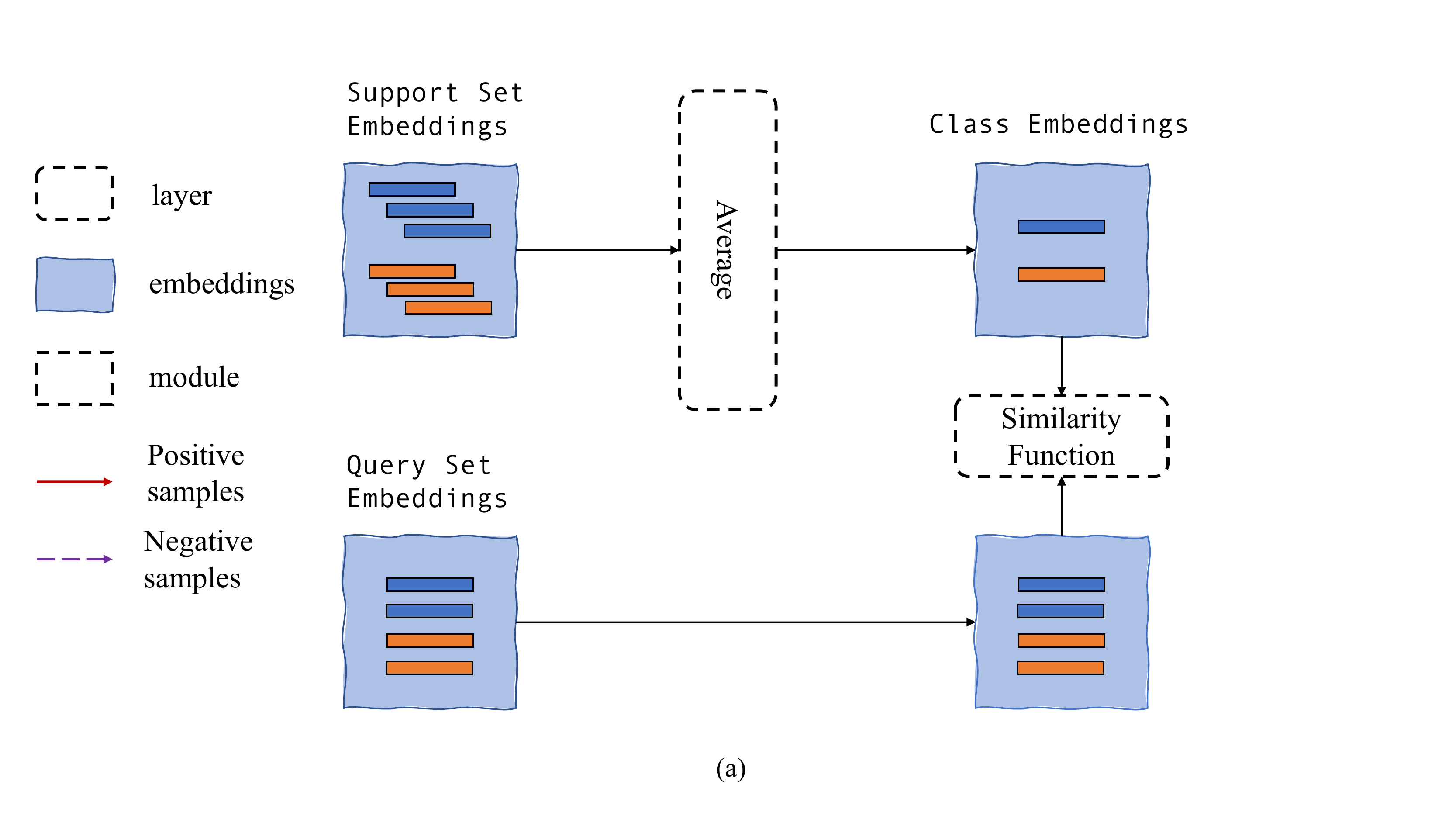}};
	\end{subfigure}
	~
	\begin{subfigure}[t]{1.0\columnwidth}
	    \draw (8,0) node[inner sep=0] {\includegraphics[width=1.0\columnwidth, height=6cm, trim={0cm 0cm 0cm 0cm}, clip]{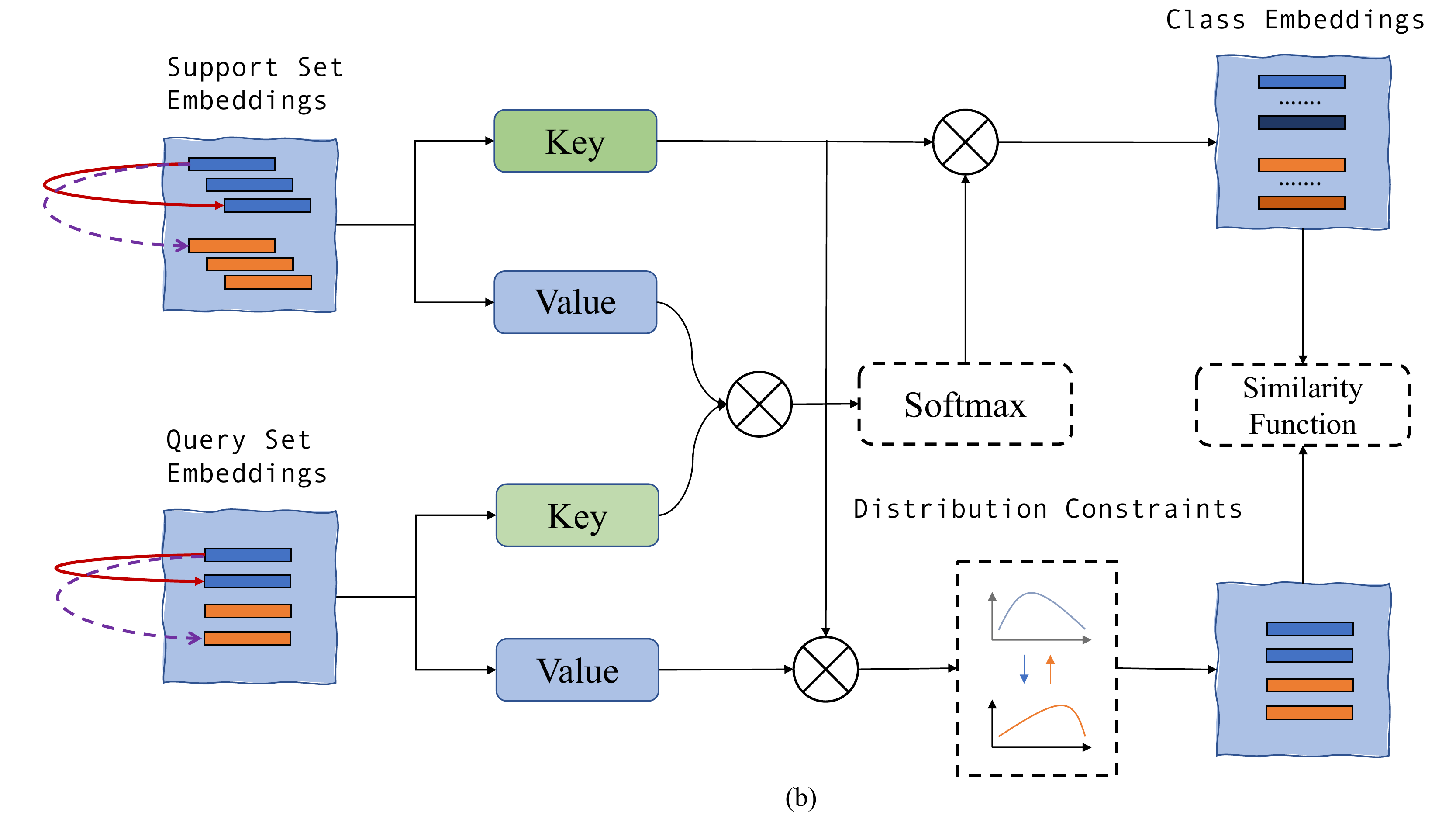}};
	\end{subfigure}

	\end{tikzpicture}
	\caption{ (a) The main architecture of ProtoNets. (b) The design of our ProtoCACL model. The positive samples (\textcolor{red}{red line}) and negative samples (\textcolor{violet}{purple dashed line}) of contrastive learning can attend to both support set and query set.  The encoder is omitted for brevity in both methods.
	}\vspace*{-4mm}
	\label{fig:framework}
\end{figure*}

\section{Methodology}

In this section, we propose a novel model, namely ProtoNet with cross-attention and contrastive learning (ProtoCACL), to address the inconsistent few-shot task. We first introduce the overall framework and then dive into each element of our ProtoCACL model, including the ProtoNet architecture, cross-attention, and contrastive learning.

\subsection{Overall Framework}

We introduce the overall framework of our proposed ProtoCACL model in detail. Compared with the main architecture of ProtoNets as shown in Figure~\ref{fig:framework}(a), our model proposes two novel parts, cross-attention and contrastive learning, as shown in Figure~\ref{fig:framework}(b).

\noindent \textbf{Prototypical Networks. } The key idea of ProtoNet is to classify sentences in the query set according to its similarity with the prototypes learned from the support set without considering the interactions between the support set and query set as shown in Figure~\ref{fig:framework}(a). In contrast, our ProtoCACL utilize ProtoNet as backbone and propose cross-attention and contrastive learning mechanisms.

\noindent \textbf{Cross-Attention. } Most of the previous works either only model the support set and query set with a shared encoder \citep{han-etal-2018-fewrel} or explicitly model the unidirectional information flow from the query to the support \citep{gao2019hybrid}. Based on ProtoNet, we introduce cross-attention mechanism to consider the relationships and cross-impact by building a bidirectional connection between the support set and the query set.

\noindent \textbf{Contrastive Learning. } We introduce the contrastive learning loss as an additional constraint to force the sentences with same relations to be close, and those with different relations separate. We add the this constraint to both the 
% three types of constraints: (1) only support set, (2) only query set and 3) both 
support set and query set. The experimental results demonstrate the cross-attention mechanism and contrastive learning constraints cooperate with each other and achieve higher accuracy together.

\subsection{Prototypical Networks}
Given the text in either the support set or query set, the ProtoNet firstly utilizes a shared encoder to encode the whole text into the feature space and obtain the text embedding $\bm{x}$. 
% We denote such encoding operation as follows:
% \begin{equation}
%     \bm{x} = f_\sigma (\bm{x})
%     ~,
% \end{equation}
After computing text embeddings, for each relation $r$, its prototype embedding $\bm{c}_r$ can be calculated by the following formula:
\begin{equation}
    \bm{c}_r = \frac{1}{n_r} \sum_{j=1}^{n_r} \bm{x}_r^j 
    ~,
\end{equation}
% where $c_i$ is computed from the feature space and represents the relation . 
where all the samples from the support set are weighted equally in vanilla ProtoNet framework. Finally, we calculate the distance between the each relation's prototype embedding and the sample vector from query set by a predefined similarity function $\mathrm{sim}(\cdot,\cdot)$:

\begin{equation}
    p_\sigma (y=r | \bm{x}) = \frac{\exp (\mathrm{sim}(\bm{x}, \bm{c}_r))}{\sum_{r' \in \mathbb{R}} \exp(\mathrm{sim}(\bm{x}, \bm{c}_{r'}))}
    ~.
\end{equation}

Here, we adopt Euclidean distance as similarity function following \cite{NIPS2017_cb8da676}, as it outperforms the other distance functions \citep{NIPS2017_cb8da676}.

\subsection{Cross-Attention}
% Given the support set with labeled samples $(\bm{s}, r)$, and the query set with only unlabeled samples $\bm{s}_q$, we use cross-attention to learn better vector representations of (1) the query and (2) the relation prototypes.

Most existing works either only model the support set and query set with a shared encoder \citep{han-etal-2018-fewrel}, or explicitly model the single information flow from the query to the support \citep{gao2019hybrid}. In contrast, we introduce a cross-attention mechanism to consider their relationships and interactions by building bidirectional attentions between the support set and the query set.

\paragraph{Support $\rightarrow$ Query. 
% Learning Query Embeddings by the Support Set.
} Recall that we have the training set consisting of a relation set $\mathbb{R}_{\mathrm{train}}$, a support set $\mathbb{S}_{\mathrm{train}}$ with labeled samples, and a query set $\mathbb{Q}_{\mathrm{train}}$ with unlabeled samples:
%The first type of cross-attention is to learn the embedding of the query set by attending to the support set.
\begin{align*}
\small
\mathbb{R}_{\mathrm{train}} = \{ & r_1, r_2, \dots, r_{N1}\}
~,\\
\mathbb{S}_{\mathrm{train}} = \{ & (\bm{s}_r^1, r), \dots, (\bm{s}_r^{K_1}, r) | \text{ for each } r \in \mathbb{R}_{\mathrm{train}} \}
~,\\
\mathbb{Q}_{\mathrm{train}} = \{ & \bm{q}^1, \dots, \bm{q}^{|\mathbb{Q}|} \}
~.
\end{align*}
% Each sentence have one corresponding relation no matter in support set or query set. 

For each labeled sample $\bm{s}_r^i$ (which is the $i$-th sample with the labeled relation type $r$) in the support set $\mathbb{S}$, we attend it with all unlabeled samples $\bm{q}_j$ in the query set $\mathbb{Q}$. Hence, for each sample $\bm{s}_r^i$, applying this attention mechanism results in a distribution $\bm{d}_r^i$:
\begin{equation}
 \bm{d}_r^i = \bm{s}_r^i \cdot [\bm{q}^T_1, \bm{q}^T_2, \dots, \bm{q}^T_{|\mathbb{Q}|}]
~,
\end{equation}
where we use the attention mechanism by dot product, and $\cdot^T$ denotes the vector transpose.

This $\bm{d}_r^i$ is a distribution over all unlabeled samples in the query set, where the unlabeled samples with the same relation as $r$ will get higher attention scores and others with relations other than $r$ will get lower attention scores.

Among the labeled samples in the support set $\mathbb{S}$, the distributions for samples with the same relation $r$, namely $\bm{d}_r^1, \dots, \bm{d}_r^{K_1}$, should be similar, whereas distributions for samples with different relations should be different. As shown in Figure~\ref{fig:class_matrix}, we first calculate a matrix which represents the distances between different distributions, and then minimize the distribution distance among samples with the same relations and maximize the distance among samples with different relations. Denote the set of all distributions $\bm{d}_r^i$ as $\mathbb{D}$. The distribution constraints can be written as follows:
\begin{figure}[t]
    \centering
    \centering
	\footnotesize
	\begin{tikzpicture}
	\draw (0,0) node[inner sep=0] {\includegraphics[width=1.0\columnwidth, height=5cm, trim={0cm 0cm 0cm 0cm}, clip]{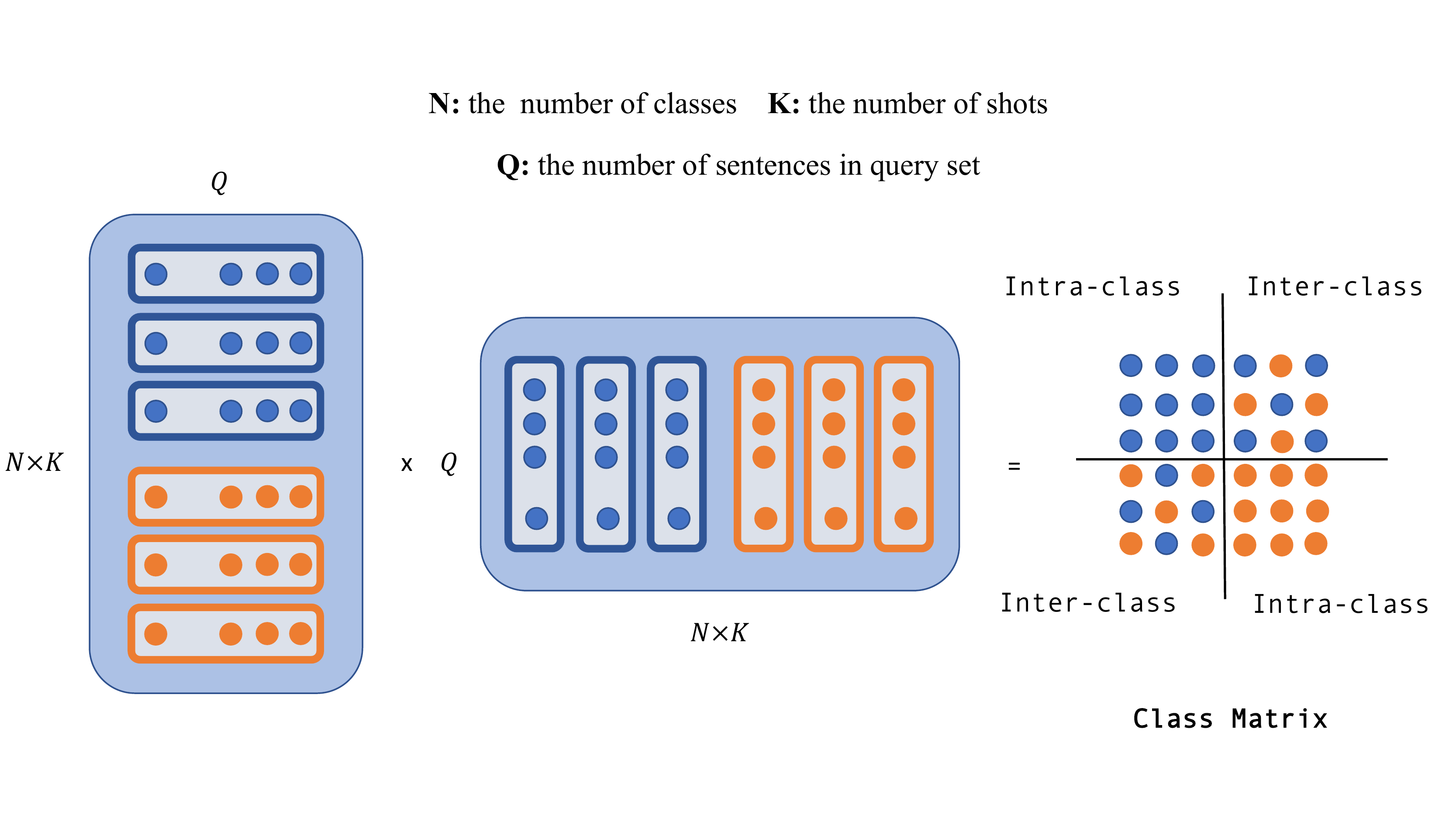}};
	\end{tikzpicture}
	\caption{An example operation of the distribution constraints by a 2-way 3-shot example. The diagonal elements in the \textit{Class Matrix} represent the intra-class variance and elements out of the diagonal represent the inter-class variance.
	}\vspace*{-4mm}
	\label{fig:class_matrix}
\end{figure}
% \begin{align}
% \mathcal{L}_{\mathrm{Dist}} &= \frac{\sum_{\bm{a},\bm{b} \in \mathbb{D}} (\bm{a} \cdot \bm{b}) \phi(\bm{a}, \bm{b})} {\sum_{\bm{a},\bm{b} \in \mathbb{D}} (\bm{a} \cdot \bm{b}) (1 - \phi(\bm{a}, \bm{b})) }
% \\
% &= \frac{\sum_{r\in \mathbb{R}} \sum_{i=0}^{K_1} \sum_{j=0}^{K_1} \bm{d}_r^i \cdot \bm{d}_r^j} {\sum_{r\in \mathbb{R}} \sum_{r' \in \mathbb{R}\setminus \{r\}} \sum_{i=0}^{K_1} \sum_{j=0}^{K_1} \bm{d}_r^i \cdot \bm{d}_{r'}^j }
% ~,
% \end{align}
\begin{align}
\mathcal{L}_{\mathrm{Dist}} &= \frac{\sum_{\bm{a},\bm{b} \in \mathbb{D}} \mathrm{dis}(\bm{a}, \bm{b}) \phi(\bm{a}, \bm{b})} {\sum_{\bm{a},\bm{b} \in \mathbb{D}} \mathrm{dis}(\bm{a}, \bm{b}) (1 - \phi(\bm{a}, \bm{b})) }
\\
&= \frac{\sum_{r\in \mathbb{R}} \sum_{i=0}^{K_1} \sum_{j=0}^{K_1} \mathrm{dis}(\bm{d}_r^i , \bm{d}_r^j)} {\sum_{r\in \mathbb{R}} \sum_{r' \in \mathbb{R}\setminus \{r\}} \sum_{i=0}^{K_1} \sum_{j=0}^{K_1} \mathrm{dis}(\bm{d}_r^i , \bm{d}_{r'}^j) }
~,
\end{align}
where $\phi(\bm{a}, \bm{b})$ is an indicator function which equals 1 if $\bm{a}$ and $\bm{b}$ are the distributions of two samples labeled with the same relation type in the support set, and $\mathrm{dis}(\cdot, \cdot)$ is the distance function 
% Here, we adopt the dot product function to calculate the distance 
between two distributions.

An alternative perspective to understand the loss formula above comes from the variance of data. The intrinsic goal of the distribution constraints is to maximize the inter-class variance and minimize intra-class variance. The inter-class variance and intra-class variance play a key role under the setting of inconsistent $K$, according to the theoretical analysis of the number of shots in few-shot learning by \citet{cao2019theoretical}.

\paragraph{Query $\rightarrow$ Support. 
% Learning Query-Specific Relation Class Embeddings.
} 
The second type of cross-attention is to learn the embedding of relation prototypes by attending to the query set. A relation prototype $\bm{c}_r$ is an overall embedding for a relation $r$ learned from the support set samples $\bm{s}_r^1, \dots, \bm{s}_r^{K_1}$ which all share the same relation $r$. In contrast to the previous practice in ProtoNets which obtains the relation prototype by simply averaging over the embeddings of all samples with the same relation, we use the query set to guide the learning of the relation prototypes followed \citep{gao2019hybrid}. Specifically, the relation prototype $\bm{c}_r$ of the relation $r$ is learned as follows:
% followed \citep{gao2019hybrid}.
\begin{align}
    \bm{c}_r &= \sum_{i=1}^{K_1} \alpha_i \bm{s}_r^i
\end{align}
where $\alpha_i$ is defined as follows,
\begin{align}
\alpha_i &= \frac{ \exp(e_i)}{\sum_{j=1}^{K_1} \exp(e_j)} \\
e_i &= \textrm{vecsum} \{ \textrm{tanh}(h(\bm{s}_r^i) \odot h([\bm{q}_1, \bm{q}_2, \dots, \bm{q}_{|\mathbb{Q}|}])) \}
\end{align}
where $h(\cdot)$ is a linear projection, $\odot$ is the element-wise product, and 
% $\textrm{tanh}(\cdot)$ is one of activate functions and 
$\textrm{vecsum}\{\cdot\}$ means the sum of all elements of the vector. In this way, samples that are closer to the given query get higher weights, and the learned prototypes can be more robust and specific over different classes as well as reduce the ambiguous samples.

\subsection{Contrastive Learning for Representation}
Due to the poorly-sampled few shots \citep{fei2020melr,ding2021prototypical}, these few samples with the same relation may disperse, and others with different relations overlap in the feature space, resulting in under-fitting prototypes. For example, there are many sentences that express the same relations, but contain different synthetic and semantic information. Also many sentences with different relations have similar semantic and synthetic information \citep{ding2021prototypical}. Here, we introduce the contrastive learning loss as additional constraint to enforce samples with the same relation to have more similar embeddings, and samples with different relations to have more dissimilar embeddings, aiming to improve the latent representation of the support and query sets.

Specifically, we define a distance function followed \cite{ding2021prototypical} first:
\begin{equation}
    \mathrm{dis}(\bm{s}_i, \bm{s}_j) = 1 / (1 + \exp{\frac{\bm{s}_i}{||\bm{s}_i||} \cdot \frac{\bm{s}_j}{||\bm{s}_j||} } )
\end{equation}
and then we define the contrastive learning loss $\mathcal{L}_{\mathrm{CL}}$ as follows: 
\begin{equation}
    \mathcal{L}_{\mathrm{CL}} = \frac{ \sum_{r\in \mathbb{R}} \sum_{i=0}^{K_1} \sum_{j=0}^{K_1} \exp{(\phi(\bm{s}_r^i, \bm{s}_r^j) \cdot \mathrm{dis}(\bm{s}_r^i, \bm{s}_r^j))}}{ \sum_{r\in \mathbb{R}} \sum_{i=0}^{K_1} \sum_{j=0}^{K_1}  \exp{\left( \left(1 - \phi(\bm{s}_r^i, \bm{s}_r^j) \right) \cdot \mathrm{dis}(\bm{s}_r^i, \bm{s}_r^j)\right)}}
\end{equation}

\begin{table*}[t]
    \centering
    \small
    \begin{tabular}{l c cccc c cccc}
    \toprule[1pt]
    \multirow{2}{*}{Model} & \multirow{2}{*}{$K_1$} &  \multicolumn{4}{c}{$N_1=N_2=5$} & &  \multicolumn{4}{c}{$N_1=N_2=10$} \\ \cline{3-6} \cline{8-11}
    & & $K_2=1$ & $K_2=5$ & $K_2=10$ & $K_2=20$ & & $K_2=1$ & $K_2=5$ & $K_2=10$ & $K_2=20$ \\
    \hline
    Proto & 5 & 73.24 \tiny $\pm$0.39 & 85.85 \tiny $\pm$0.16 & 88.08 \tiny $\pm$0.16 & 89.25 \tiny $\pm$0.23 & & 60.36 \tiny $\pm$0.49 & 75.67 \tiny $\pm$0.29 & 78.91 \tiny $\pm$0.32 & 80.78 \tiny $\pm$0.20 \\
    ProtoCACL & 5 & \textbf{73.33} \tiny $\pm$0.27 & \textbf{86.44} \tiny $\pm$0.10 & \textbf{88.32} \tiny $\pm$0.10 & \textbf{89.50} \tiny $\pm$0.13 & & \textbf{60.42} \tiny $\pm$0.34 & \textbf{76.40} \tiny $\pm$0.12 & \textbf{79.36} \tiny $\pm$0.28 & \textbf{81.15} \tiny $\pm$0.10 \\
    \hline
    Proto & 10 & 72.53 \tiny $\pm$0.21 & 85.73 \tiny $\pm$0.16 & 87.85 \tiny $\pm$0.13 & 89.13 \tiny $\pm$0.10 & & 59.30 \tiny $\pm$0.09 & 75.39 \tiny $\pm$0.17 & 78.52 \tiny $\pm$0.33 & 80.51 \tiny $\pm$0.19 \\
    ProtoCACL & 10 & \textbf{72.71} \tiny $\pm$0.15 & \textbf{86.17} \tiny $\pm$0.18 & \textbf{88.45} \tiny $\pm$0.16 & \textbf{89.64} \tiny $\pm$0.15 & & \textbf{60.24} \tiny $\pm$0.30 & \textbf{76.19} \tiny $\pm$0.13 & \textbf{79.55} \tiny $\pm$0.14 & \textbf{81.41} \tiny $\pm$0.13 \\
    \hline
    Proto & 20 & 71.41 \tiny $\pm$0.53 & 85.53 \tiny $\pm$0.21 & 87.57 \tiny $\pm$0.13 & 88.90 \tiny $\pm$0.20 & & 58.28 \tiny $\pm$0.28 & 75.12 \tiny $\pm$0.25 & 78.32 \tiny $\pm$0.06 & 80.30 \tiny $\pm$0.09 \\
    ProtoCACL & 20 & \textbf{71.94} \tiny $\pm$0.28 & \textbf{85.88} \tiny $\pm$0.09 & \textbf{88.18} \tiny $\pm$0.13 & \textbf{89.65} \tiny $\pm$0.15 & & \textbf{59.69} \tiny $\pm$0.09 & \textbf{75.93} \tiny $\pm$0.24 & \textbf{79.48} \tiny $\pm$0.24 & \textbf{81.55} \tiny $\pm$0.18 \\
    \bottomrule[1pt]
    \end{tabular}
    \caption{Accuracy comparison between ProtoNets with or without the CACL under inconsistent $K$ setting ($N_1$-$N_2$-trainK-testK) as way is set to 5 or 10 respectively during training and testing.}
    \label{tab:inconsistent_k_5}
\end{table*}
Note that the distribution of relation types is uniform in the support set, but unknown in the query set, which can inﬂuence the performance of contrastive learning loss.

\begin{comment}
\begin{align}
    \mathcal{L}_{\mathrm{CL}} &= - \frac{\sum_{i,j} (x_i \cdot x_j) \phi(x_i, x_j)} {\sum_{i,j} (x_i \cdot x_j) (1 - \phi(x_i, x_j)) }
\\
\begin{split}
    \mathcal{L}_{\mathrm{CL}} &= - \mathbb{E} [\phi(a, b) \mathrm{Distance}(a, b) + 
\\
& (1 - \phi(a, b)) \max(0, \mathrm{margin} - \mathrm{Distance}(a, b))]
~.
\end{split}
% \mathcal{L}_{\mathrm{Dist}} &= \frac{\sum_{\bm{a},\bm{b} \in \mathbb{D}} (\bm{a} \cdot \bm{b}) \phi(\bm{a}, \bm{b})} {\sum_{\bm{a},\bm{b} \in \mathbb{D}} (\bm{a} \cdot \bm{b}) (1 - \phi(\bm{a}, \bm{b})) }
\end{align}
\end{comment}

\subsection{Training Objective}
The overall training objective consists of three parts: (1) cross entropy loss of original ProtoNet, (2) cross-attention loss, and (3) contrastive learning loss. Namely,
\begin{align}
    \mathcal{L} = \lambda_{\mathrm{CE}} \mathcal{L}_{\mathrm{CE} }+ \lambda_{\mathrm{Dist}} \mathcal{L}_{\mathrm{Dist}} + \lambda_{\mathrm{CL}} \mathcal{L}_{\mathrm{CL}}
    ~,
\end{align}
where $\lambda_{\mathrm{CE}}$, $\lambda_{\mathrm{Dist}}$ and $\lambda_{\mathrm{CL}}$ are the hyperparameters that represent the weights of the three losses.

\section{Experiments and Results}

This section will show the performance of our proposed ProtoCACL under various inconsistent few-shot RC settings.

\subsection{Dataset and Evaluation Metrics}
We choose the commonly used few-shot RC dataset, FewRel \citep{han-etal-2018-fewrel},\footnote{https://github.com/thunlp/FewRel} as our benchmark. The dataset consists of 100 relations and each relation has 700 sentences. The training, validation, and test sets have 64, 20, 16 relations, respectively. Only the training set and validation set of FewRel are released publicly, and there is no overlapping relations between training and validation set. Due to the unavailability of the test set, we conduct our experiments and performance analysis on the validation set. Following the practice of \citet{gao2019hybrid}, we use accuracy as the evaluation metric.

\subsection{Implementation Details}
Following \citet{gao2019hybrid}, we adopt CNN \citep{lai2015recurrent} to encode all tokens in the sentence and use GloVe \citep{pennington2014glove} as pre-trained word embeddings. Instead of simply feeding more classes, we design several inconsistent few-shot settings and conduct multiple experiments. For inconsistent $K$ problem, we fix the value of $N$ as 5 or 10 and set training shots as [5, 10, 20], testing shots as [1, 5, 10, 20]. For inconsistent $N$ problem, we fix the value of $K$ as 5 or 10 and set training ways as [5, 10, 20], testing ways as [5, 10]. Note that setting testing ways as 1 is meaningless and the validation set only contain 16 relations. Thus we do not set testing ways as 1 or 20. To make a fair comparison, we train ProtoNet and ProtoCACL in the same way with 30K iterations and test 10K iterations.

\subsection{Performance under Inconsistent K}

% 三个角度，参考ppt
Table~\ref{tab:inconsistent_k_5} shows the performance of inconsistent $K$ problems under 5-way and 10-way setting. (1) From left to right, when the training shot is fixed, more testing shots brings more improvement especially from 1-shot to 5-shot, regardless the method. (2) From top to bottom, when the testing shot is fixed, more training shots damage the performance in ProtoNet. The overlap of additional training samples in feature space make it hard to learn independent prototype and introduce more noises. Our CACL method alleviate this problem by introducing cross-attention and contrastive learning constraints especially at 10-shot and 20-shot. (3) From the top left to bottom right, when the training shot and testing shot are same (i.e., standard few-shot setting), more shots brings more improvements and our CACL can further improve it compared with ProtoNet (e.g. 81.55 - 79.55 vs 80.30 - 78.52 under 10-way).

% 对角线取差值画一个图

% \begin{table}[h]
%     \centering
%     \begin{tabular}{l c cc}
%     \toprule[1pt]
%     \multirow{2}{*}{Model} & Training &  \multicolumn{2}{c}{Testing Ways} \\ \cline{3-4} & ways & 5 & 10 \\
%     \hline
%     Proto & 5 & 85.85 \tiny $\pm$ 0.16 & 75.28 \tiny $\pm$ 0.27   \\
%     ProtoCACL & 5 & \textbf{86.44} \tiny $\pm$ 0.10 & \textbf{76.11} \tiny $\pm$ 0.19   \\
%     \hline
%     Proto & 10 & 86.10 \tiny $\pm$ 0.15 & 75.67 \tiny $\pm$ 0.29  \\
%     ProtoCACL & 10 & \textbf{86.48} \tiny $\pm$ 0.18 & \textbf{76.40} \tiny $\pm$ 0.12  \\
%     \hline
%     Proto & 20 & 86.22 \tiny $\pm$ 0.23 & 75.96 \tiny $\pm$ 0.38  \\
%     ProtoCACL & 20 & \textbf{86.64} \tiny $\pm$ 0.08 & \textbf{76.32} \tiny $\pm$ 0.13 \\
%     \bottomrule[1pt]
%     \end{tabular}
%     \caption{Accuracy comparison between ProtoNets with or without the CACL under inconsistent $N$ setting (trainN-testN-5-5) as shot is set to 5 during training and testing.}
%     \label{tab:inconsistent_n_5}
% \end{table}

\begin{table}[t]
    \centering
    \setlength{\tabcolsep}{3pt}
    \small
    \begin{tabular}{l c cc c cc}
    \toprule[1pt]
    \multirow{2}{*}{Model} & \multirow{2}{*}{$N_1$} &  \multicolumn{2}{c}{$K_1=K_2=5$} & &  \multicolumn{2}{c}{$K_1=K_2=10$} \\ \cline{3-4}\cline{6-7} & & $N_2=5$ & $N_2=10$ & & $N_2=5$ & $N_2=10$\\
    \hline
    Proto & 5 & 85.85 \tiny $\pm$0.16 & 75.28 \tiny $\pm$0.27  & & 87.85 \tiny $\pm$0.13 & 78.23 \tiny $\pm$0.23 \\
    ProtoCACL & 5 & \textbf{86.44} \tiny $\pm$0.10 & \textbf{76.11} \tiny $\pm$0.19  & & \textbf{88.45} \tiny $\pm$0.16 & \textbf{78.94} \tiny $\pm$0.12 \\
    \hline
    Proto & 10 & 86.10 \tiny $\pm$0.15 & 75.67 \tiny $\pm$0.29 & & 88.14 \tiny $\pm$0.10 & 78.52 \tiny $\pm$0.33 \\
    ProtoCACL & 10 & \textbf{86.48} \tiny $\pm$0.18 & \textbf{76.40} \tiny $\pm$0.12 & & \textbf{88.68} \tiny $\pm$0.08 & \textbf{79.55} \tiny $\pm$0.14\\
    \hline
    Proto & 20 & 86.22 \tiny $\pm$0.23 & 75.96 \tiny $\pm$0.38 & & 88.20 \tiny $\pm$0.14 & 78.95 \tiny $\pm$0.23  \\
    ProtoCACL & 20 & \textbf{86.64} \tiny $\pm$0.08 & \textbf{76.52} \tiny $\pm$0.17& & \textbf{88.88} \tiny $\pm$0.19 & \textbf{79.82} \tiny $\pm$0.33 \\
    \bottomrule[1pt]
    \end{tabular}
    \caption{Accuracy comparison between ProtoNets with or without the CACL under inconsistent $N$ setting (trainN-testN-$K_1$-$K_2$) as shot is set to 5 or 10 respectively during training and testing.}
    \label{tab:inconsistent_n_10}
\end{table}

\subsection{Performance under Inconsistent N}
Table~\ref{tab:inconsistent_n_10} shows the performance of inconsistent $N$ problems under 5-shot and 10-shot setting. There are two conclusions drawn from the experimental results: (1) when the training way is fixed, more testing way damage the performance. (2) when the testing way is fixed, feeding more classes during training leads to better results, which is consistent with \citep{gao2019hybrid}. Note that although more training ways increase the accuracy, the improvement is less than 1 percent but the computational cost is high.

\begin{figure}[t]
\centering
\begin{minipage}{.5\columnwidth}
    \centering
    \includegraphics[width=\columnwidth, height=0.8\columnwidth]{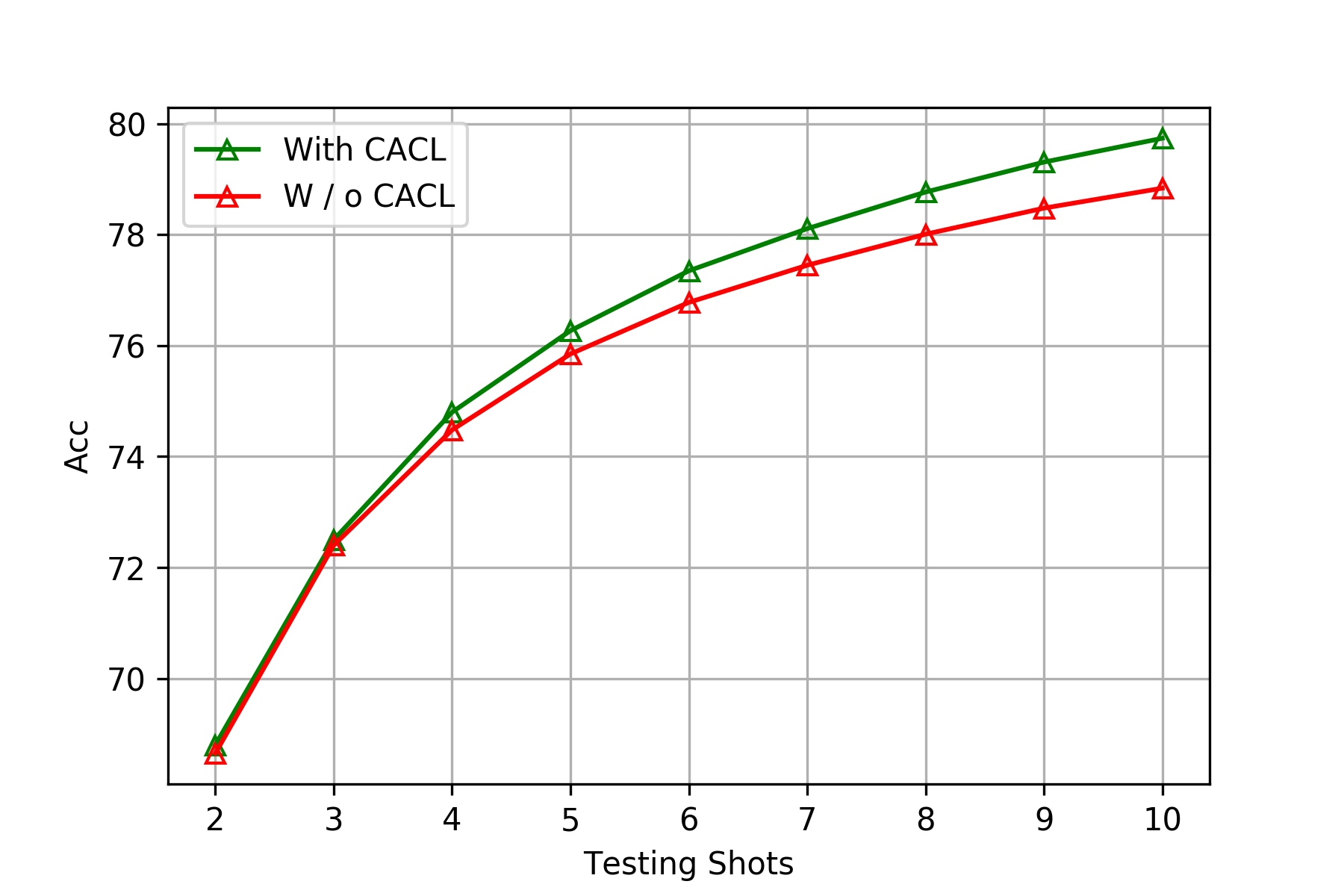} \\ (a)
\end{minipage}%
\hfill
\begin{minipage}{.5\columnwidth}
    \centering
    \includegraphics[width=\columnwidth, height=0.8\columnwidth]{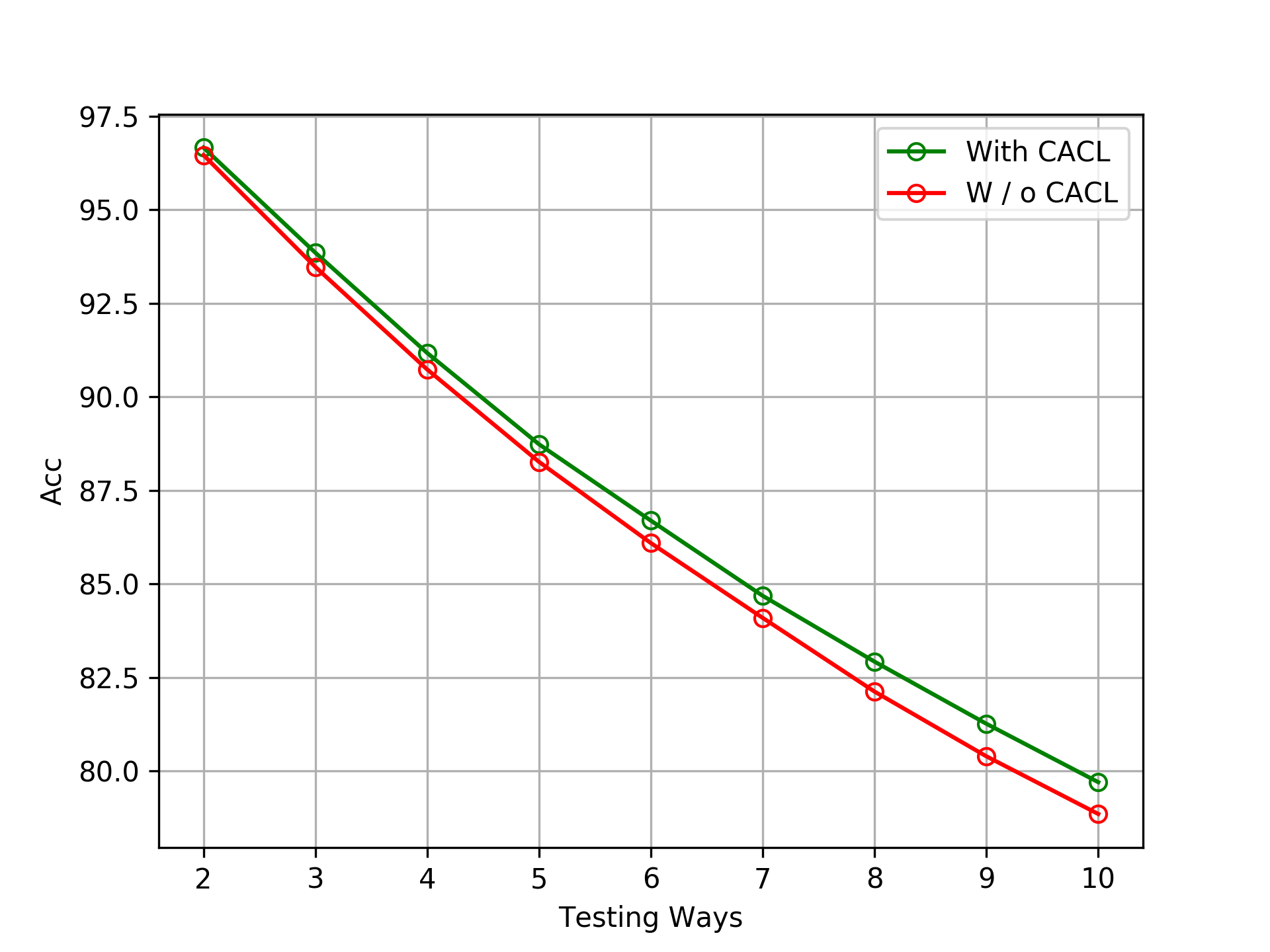}\\ (b)
\end{minipage}
\caption{Accuracy of different models under \textit{inconsistent testing shots} and \textit{inconsistent testing ways}, using trained models under standard 10-way 10-shot setting.}
\label{fig:fewer_exp}
\end{figure}

\section{Analysis and Discussion}

\subsection{Few``er''-Shot under Inconsistent Setting}
We further evaluate our CACL method under less than 10-shot or 10-way scenario. Specifically, we firstly choose two trained models under standard 10-way 10-shot setting using CACL or ProtoNet respectively and evaluate the corresponding performance under inconsistent $K$ and $N$ scenarios. As shown in Figure~\ref{fig:fewer_exp} (a) and Figure~\ref{fig:fewer_exp} (b), our method consistently outperforms the ProtoNet under both inconsistent $K$ and inconsistent $N$ settings (i.e., w/o CACL in the figures). Additionally, as the testing shot or testing way rises, the advantages of our proposed models become more obvious. We also find that the performance of the models are more sensitive under inconsistent $N$ compared with inconsistent $K$. For example, when the testing shot changes from 5-shot to 10-shot, the performance increases from about 76\% to 79\% ($\approx$ 3\% gap). However, as the testing ways keep the same adjustment, the performance drops from about 88\% to 78\% ($\approx$ 10\% gap).

\subsection{The Effects of CA or CL}

\begin{table}[t]
    \centering
    \small
    \begin{tabular}{l cc}
    \toprule[1pt]
    \textbf{Model} & \textbf{5 Way 5 Shot} & \textbf{10 Way 5 Shot} \\
    \hline
    Proto & 85.85 \tiny $\pm$0.16 & 75.67 \tiny $\pm$0.29 \\
    \hline
    Proto - S & 86.32 \tiny $\pm$0.29 & 76.10 \tiny $\pm$0.30  \\
    Proto - Q & 86.29 \tiny $\pm$0.13 & 76.18 \tiny $\pm$0.12 \\
    Proto - S\_and\_Q & 86.39 \tiny $\pm$0.34 & 76.36 \tiny $\pm$0.29 \\
    \hline
    W/o CL & 86.17 \tiny $\pm$0.20 & 76.21 \tiny $\pm$0.18 \\
    ProtoCACL & \textbf{86.44} \tiny $\pm$0.10 & \textbf{76.40} \tiny $\pm$0.12 \\
    \bottomrule[1pt]
    \end{tabular}
    \caption{Accuracy comparison between ProtoNets with or without the CACL. 10-shot, with CNN as encoder}
    \label{tab:aba_study}
\end{table}

Table~\ref{tab:aba_study} shows the ablation study of our proposed CACL method at two standard few-shot settings: 5-way 5-shot and 10-way 5-shot. ``Proto-S", ``Proto-Q" and ``Proto -S\_and\_Q" in the table represent three different types of contrastive learning constraints respectively (without cross-attention). It is obvious that either cross-attention (CA) alone or contrastive learning (CL) alone can bring improvement and their combination reaches the highest accuracy (i.e., 86.44 and 76.40). In detail, CL added to both support set and query set brings higher improvement than support only or query only. The unknown distribution in the query set also brings variance of performance at ``Proto-Q". For example,  the accuracy of ``Proto-Q" is lower than ``Proto-S" at standard 5-way 5-shot setting but higher at 10-way 5-shot setting.

\subsection{Discussion}

It is obvious that the model under inconsistent few-shot setting achieves better performance than standard setting with carefully selected $N$ and $K$ from the experimental results. In this section, we provide some suggestions to answer a practical question: \textbf{Which setting brings better improvement for inconsistent few-shot RC?}

\begin{itemize}
    \item  \textbf{Less Training Shots, or More Testing Shots.} When the testing shot is fixed, reducing training shots accordingly improves the performance. When the training shot is fixed, more testing shots help the model learn a more robust and accurate prototype and leads to higher accuracy as shown in Table~\ref{tab:inconsistent_k_5}. Compared with the standard few-shot setting, less few training shots further reduce the amount of labeled data and decrease the cost in both human annotations and computing resources.
    
    \item  \textbf{More Training Ways, or Less Testing Ways.}  As reported by \cite{gao2019hybrid}, more training ways lead to higher performance once the testing way is locked. At the same time, more testing ways propose a higher requirement and challenge for the model to learn separated and solid prototypes in the feature space, resulting in reduction in performance. It is worth considering splitting big $N$ to small $N$ during testing since the operation does not require additional labeled data and can be implemented easily.
    
    \item \textbf{Ways Matters More Than Shots.} Compare Table~\ref{tab:inconsistent_k_5} with Table~\ref{tab:inconsistent_n_10}, when the ways grow from 5-way to 10-way at each inconsistent $K$ scenario, the model's accuracy decreases by roughly 10\%; nevertheless, when it comes to inconsistent $N$ scenario, it only increases by 3\% when the ways expand from 5-shot to 10-shot. Note that a similar phenomenon is found when the shots and ways are less than 10 (see Section Few"er"-shot under inconsistent setting). The prototype of different classes contains specific features and locates in decentralized and diverse space, but the prototype of each class is still at the same place no matter the number of corresponding shots. The differences between "less-shot" prototype and "more-shot" prototype are hard to distinguish compared with inconsistent ways scenarios.
\end{itemize}

\section{Conclusion and Future Work}
In this paper, we first evaluate the performance of the ProtoNet under inconsistent few-shot settings, including inconsistent $N$ and inconsistent $K$. We find the performance is significantly affected by the inconsistency of $N$ and $K$ but previous work rarely investigate this phenomenon. And then we propose a cross attention contrastive learning framework (ProtoCACL) based on pilot experiments and previous theoretical analysis \citep{cao2019theoretical}. In detail, our CACL framework considers the mutual interaction between the support set and query set in the $N$-way $K$-shot problem and introduces some necessary constraints to model objective, leading to better representations and higher performance. Experimental results demonstrate that our methods achieve state-of-the-art results for both inconsistent $K$ and inconsistent $N$ problems. In addition, we provide some suggestions for inconsistent problems based on our experiments. In the future, we will explore more tasks under inconsistent few-shot settings and adopt more powerful encoders to further improve the performance.

% \section{ Acknowledgments}

% \bigskip
% \noindent Thank you for reading these instructions carefully. We look forward to receiving your electronic files!

\bibliographystyle{aaai}
\bibliography{ref.bib}

\end{document}